\documentclass[submission,copyright,creativecommons]{eptcs}
\providecommand{\event}{SLPCS 2016} 
\usepackage{breakurl}             

\usepackage{amsmath}
\usepackage{amssymb}  
\usepackage{latexsym} 
\usepackage{multirow}

\usepackage{tikz}
\usepackage{pgfplots}
\usetikzlibrary{trees}
\usetikzlibrary{topaths}
\usetikzlibrary{decorations.pathmorphing}
\usetikzlibrary{decorations.markings}
\usetikzlibrary{matrix,backgrounds,folding}
\usetikzlibrary{chains,scopes,positioning,fit}
\usetikzlibrary{arrows,shadows}
\usetikzlibrary{calc} 
\usetikzlibrary{chains}
\usetikzlibrary{shapes,shapes.geometric,shapes.misc}

\pgfdeclarelayer{edgelayer}
\pgfdeclarelayer{nodelayer}
\pgfsetlayers{background,edgelayer,nodelayer,main}

\tikzstyle{dot}=[circle,fill=black,draw=black]
\tikzstyle{every picture}=[baseline=(current bounding box).east,scale=0.5,node distance=5mm]
\tikzstyle{none}=[inner sep=0pt]
\tikzstyle{every loop}=[]

\tikzstyle{(null)}=[]
\tikzstyle{plain}=[]

\newcommand{\semantics}[1]{[\![ #1 ]\!]} 

\newcommand{\ovl}{\overline} 

\newcommand{\Rel}{\mathrm{Rel}}
\newcommand{\FdVect}{\mathrm{FdVect}}

\newcommand{\ket}[1]{|{#1}\rangle}

\newcommand{\relto}{\nrightarrow}

\title{ Quantifier Scope\\in  Categorical Compositional Distributional  Semantics}
\author{Mehrnoosh Sadrzadeh
\institute{School of Electronic Engineering and Computer Science, Queen Mary University London\thanks{The author thanks EPSRC for Career Acceleration Fellowship  EP/J002607/1,  the anonymous referees for their generous and useful comments, and Michael Moortgat for discussions and for directing her to  this question.}}
\email{\quad mehrnoosh.sadrzadeh@qmul.ac.uk}
}
\def\titlerunning{Quantifier Scope in  Categorical Compositional Distributional  Semantics}
\def\authorrunning{M. Sadrzadeh}
\begin{document}
\maketitle

\begin{abstract}
In  previous  work with J. Hedges, we formalised a generalised quantifiers theory of natural language in  categorical compositional distributional semantics  with the help of bialgebras. In this paper, we show how  quantifier scope ambiguity can be represented in that setting and how this representation can be generalised to branching  quantifiers. 
\end{abstract}

\section{Introduction}
Categorical Compositional Distributional semantics (CCDS) adds compositionality to  distributional semantics via a functorial passage from the syntax to the semantics of natural language \cite{CoeckeSadrClark}. Both  the syntax and the semantics are represented by  compact closed categories. The  claim is that regardless of how complex the structure of a sentence can be and what bizarre forms the words therein can take, as long as the sentence is represented in the syntax, CCDS will prescribe a vector to represent  its semantics. In practice, however, one needs more than a syntax and a functorial passage. Semantic postulates for  meanings of  words are a requirement. As long as these are of good quality,  CCDS will  provide a good theoreticalglue.  Various distributional semantics can be trusted in their ability to produce vectors for atomic words; functional words, on the other hand, need a special treatments.

In our path to provide special treatments for  functional words of natural language, in \cite{PrellerSadr} we dealt with negation,  in \cite{kartsaklis2012,RelPronMoL} we showed how meanings of verbs and relative pronouns can be represented using Frobenius algebras. Recently, we  used bialgebras to represent meanings of quantifiers \cite{HedgesSadr}. Therein,  we also developed  a relational instantiation for the model and proved that it  is sound and complete with regard to the generalised quantifier theory of  natural language by Barwise and Cooper \cite{BarwiseCooper81}.  A natural question that arises in every quantifier theory is that of scope.  The scope and branching of generalised quantifiers have been discussed in a wealth of literature, e.g.  see the collection of papers  by G\"ardenfors  \cite{Gardenfors87}, the paper by Westerstahl  in that collection  \cite{Westerstahl87}, and of course the original work of Barwise himself \cite{Barwise79}. In this short account, we show how one may deal with  scope and branching in our bialgebraic CCDS model of quantifiers.

\section{Compact Closed Categories, Bialgebras, Examples, Diagrams}
Recall that a non-symmetric compact closed category, $\cal C$, has objects $A, B$; morphisms $f \colon A
\to B$; and a monoidal tensor $A \otimes B$ that has a unit $I$, that is we have $A \otimes I \cong I \otimes A \cong A$. Furthermore, for
each object $A$ there are two objects $A^r$ and $A^l$ and  the
following morphisms:
\begin{align*}
A \otimes A^r \stackrel{\epsilon^r} {\longrightarrow} \; &I
\stackrel{\eta^r}{\longrightarrow} A^r \otimes A \hspace{1cm}
A^l \otimes A \stackrel{\epsilon^l}{\longrightarrow} \; I
\stackrel{\eta^l}{\longrightarrow} A \otimes A^l\
\end{align*}
These morphisms satisfy four equalities, known by `yanking', which we will not give here and the reader can check them here \cite{CoeckeSadrClark}.

A bialgebra  in a  symmetric monoidal  category $({\cal
  C}, \otimes, I, \sigma)$ is a tuple $(X,  \delta, \iota, \mu, \zeta)$ where,
for $X$ an object of ${\cal C}$, the triple $(X, \delta, \iota)$ is  an internal comonoid and  $(X, \mu, \zeta)$ is  an internal  monoid. This means that we have  the following morphisms in $\cal C$:
\[
\delta \colon X \to X \otimes X \quad \iota \colon X \to I\qquad
\mu \colon  X \otimes X \to X  \quad \zeta \colon I \to X
\]
where $\delta$ and $\iota$ are  coassociative and counital    and $\mu$ and $\zeta$ are associative and unital.  $\delta$ and $\mu$ satisfy the following  four equations \cite{McCurdy}:
\begin{align*}
\iota \circ \mu &= \iota \otimes \iota		&&\text{(Q1)} \\
\delta \circ \zeta &= \zeta \otimes \zeta	&&\text{(Q2)} \\
\delta \circ \mu &= (\mu \otimes \mu) \circ (\operatorname{id}_X \otimes \sigma_{X,X} \otimes \operatorname{id}_X) \circ (\delta \otimes \delta) && \text{(Q3)} \\
\iota \circ \zeta &= \operatorname{id}_I	&&\text{(Q4)}
\end{align*}

Among examples of compact closed categories are category of sets and relations $\Rel$ and category of finite dimensional vector spaces and linear maps $\FdVect$. Both $\Rel$ and $\FdVect$  are symmetric. Hence in these categories for any two objects $A,B$ we have $A \otimes B \cong B \otimes A$.   As a result, for any object $A$ we obtain that $A^r = A^l = A^*$, for $A^*$ a dual object. In $\Rel$,  for any set $S$, we have that $S^* \cong S$. In $\FdVect$, this holds for vector spaces that have a fixed basis.  In $\Rel$, given a set $S$ with elements $s_i, s_j \in S$,  the epsilon and eta maps are given as follows:

\vspace{-0.5cm}
\begin{align*}
\epsilon = \epsilon^l  =  \epsilon^r \colon   S \times S \nrightarrow I & \qquad \qquad
(s_i, s_j) \epsilon \star \iff s_i = s_j \\
\eta = \eta^l = \eta^r \colon   I  \relto S \times S &  \qquad \qquad
\star \eta (s_i, s_j) \iff s_i = s_j
\end{align*}

\noindent
Here, $\times$ is the tensor of the category with the singleton set $I = \{\star\}$  as its unit. 
 In $\FdVect$,  given a basis $\{r_i\}_i$ for a vector space $V$, the epsilon and eta maps  are  as follows:

\vspace{-0.5cm}
\begin{align*}
\epsilon = \epsilon^l  =  \epsilon^r \colon   V \otimes V \to \mathbb{R}&\qquad  \qquad
\epsilon\left(\sum_{ij} c_{ij} \ (\psi_i \otimes \phi_j) \right) =  \sum_{ij} c_{ij} \langle \psi_i \mid \phi_j \rangle\\
\eta = \eta^l = \eta^r \colon   \mathbb{R} \to V \otimes V &\qquad \qquad
\eta\left(1 \right) =  \sum_i (\ket{r_i} \otimes \ket{r_i})
\end{align*}
\noindent
Here $\otimes$ is the tensor product of two vector spaces with  $\mathbb{R}$ as its unit. 

As it will become apparent  in the next section,   generalised quantifiers  are maps of the form ${\cal P}(U) \to {\cal P}{\cal P}(U)$, for $U$ a  universe of reference. Thus, the  bialgebra structures that we are interested in are on powerset objects. In $\Rel$, these are over objects of the form ${\cal P}(U)$. In $\FdVect$, they are over vector spaces spanned by such objects, that is $V_{{\cal P}(U)}$. In  $\Rel$, for $A, B, C \subseteq U$,  they are defined as follows:

\vspace{-0.5cm}
\begin{align*}
\delta &\colon   {\cal P}(U)\relto {\cal P}(U) \times {\cal P}(U) \quad &\quad
A \delta (B, C) \iff A = B = C \\
\iota& \colon {\cal P}(U) \relto  \{\star\}   \quad & \quad A \iota \star \iff \text{ (always true)} \\
\mu & \colon  {\cal P}(U) \times{\cal P}(U) \relto{\cal P}(U)  \quad &\quad 
(A, B) \mu C \iff A \cap B = C \\
\zeta& \colon  \{\star\}  \relto{\cal P}(U)  \quad &\quad \star \zeta A \iff A = U
\end{align*}

\noindent
The coalgebraic $\delta$ map copies its input $A$ into its two outputs $B$ and $C$, so we have $A = B = C$.  Its unit $\iota$ relates any subset $A$ of $U$ to the single element in $\{\star\}$.  The algebraic $\mu$ map, takes two subsets $A$ and $B$ and relates them to their intersection $A \cap B$. Its unit $\zeta$ relates  the $\star$ to the universe of reference.  The abov definitions become as follows  in $\FdVect$:
\begin{align*}
\delta &\colon   V_{{\cal P}(U)} \to  V_{{\cal P}(U)} \otimes V_{{\cal P}(U)} \quad &\quad  \delta \ket A &= \ket A \otimes \ket A \\
\iota& \colon V_{{\cal P}(U)} \to  \mathbb{R}   \quad & \quad  \iota \ket A &= 1 \\
\mu & \colon   V_{{\cal P}(U)} \times V_{{\cal P}(U)} \to V_{{\cal P}(U)} \quad &\quad  \mu (\ket A \otimes \ket B) &= \ket{A \cap B}\\
\zeta& \colon  \mathbb{R}  \to V_{{\cal P}(U)},  \quad &\quad \zeta &= \ket U \\
\end{align*}

\vspace{-0.5cm}
\noindent
where they have the same functionalities.  $\delta$ copies its input basis vector and $\iota$ relates it to $1 \in \mathbb{R}$.  $\mu$ sends its two input basis vectors  to the basis vector obtained by taking the intersection of them. Recall that the basis vectors are subsets of $U$, hence their intersection  is set intersection.   In previous work \cite{HedgesSadr},  we showed how these definitions satisfy the bialgebra conditions (Q1) to (Q4).

Diagrammatically,   a morphism $f \colon A \to B$ and an identity on an object $A$ of a compact closed category are depicted as follows:

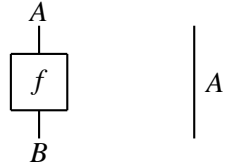
\begin{center}
  {%
\beginpgfgraphicnamed{compact-diag}
\begin{tikzpicture}[scale=0.75]
	\begin{pgfonlayer}{nodelayer}
		\node [style=none] (0) at (-9, -1) {};
		\node [style=none] (1) at (-7, -1) {};
		\node [style=none] (2) at (-7, 1) {};
		\node [style=none] (3) at (-9, 1) {};
		\node [style=none] (4) at (-8, 0) {$f$};
		\node [style=none] (5) at (-8, 1) {};
		\node [style=none] (6) at (-8, 2) {};
		\node [style=none] (7) at (-8, -1) {};
		\node [style=none] (8) at (-8, -2) {};
		\node [style=none] (9) at (-8, 2.5) {$A$};
		\node [style=none] (10) at (-8, -2.5) {$B$};
		\node [style=none] (11) at (-2.5, 2) {};
		\node [style=none] (12) at (-2.5, -2) {};
		\node [style=none] (13) at (-1.75, 0) {$A$};
	\end{pgfonlayer}
	\begin{pgfonlayer}{edgelayer}
		\draw [style=thick] (3.center) to (0.center);
		\draw [style=thick] (3.center) to (2.center);
		\draw [style=thick] (2.center) to (1.center);
		\draw [style=thick] (1.center) to (0.center);
		\draw [style=thick] (6.center) to (5.center);
		\draw [style=thick] (7.center) to (8.center);
		\draw [style=thick] (11.center) to (12.center);
	\end{pgfonlayer}
\end{tikzpicture}}
\endpgfgraphicnamed}
\end{center}

In concrete categories, morphisms of the form $I \to A$ represent elements of $A$. These are depicted as follows for elements of $A, A \otimes B$, and $A \otimes B \otimes C$; elements of other tensor objects are depicted similiarly:

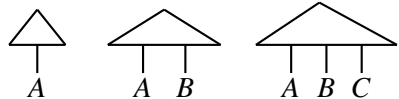
\begin{center}
  {%
\beginpgfgraphicnamed{compact-diag-triangle}
\begin{tikzpicture}[scale=0.75]
	\begin{pgfonlayer}{nodelayer}
		\node [style=none] (0) at (-5, 0) {};
		\node [style=none] (1) at (-1, 0) {};
		\node [style=none] (2) at (-3, 1.25) {};
		\node [style=none] (3) at (-3.75, 0) {};
		\node [style=none] (4) at (-3.75, -1) {};
		\node [style=none] (5) at (-3.75, -1.5) {$A$};
		\node [style=none] (6) at (-2.25, 0) {};
		\node [style=none] (7) at (-2.25, -1.5) {$B$};
		\node [style=none] (8) at (-2.25, -1) {};
		\node [style=none] (9) at (2.75, -1) {};
		\node [style=none] (10) at (2.75, -1.5) {$B$};
		\node [style=none] (11) at (2.5, 1.5) {};
		\node [style=none] (12) at (1.5, 0) {};
		\node [style=none] (13) at (5.25, 0) {};
		\node [style=none] (14) at (1.5, -1.5) {$A$};
		\node [style=none] (15) at (1.5, -1) {};
		\node [style=none] (16) at (0.25, 0) {};
		\node [style=none] (17) at (2.75, 0) {};
		\node [style=none] (18) at (4, -1.5) {$C$};
		\node [style=none] (19) at (4, 0) {};
		\node [style=none] (20) at (4, -1) {};
		\node [style=none] (21) at (-8.5, 0) {};
		\node [style=none] (22) at (-7.5, -1.5) {$A$};
		\node [style=none] (23) at (-7.5, 0) {};
		\node [style=none] (24) at (-6.5, 0) {};
		\node [style=none] (25) at (-7.5, -1) {};
		\node [style=none] (26) at (-7.5, 1.25) {};
		\node [style=none] (29) at (2.5, 0.5) {};
	\end{pgfonlayer}
	\begin{pgfonlayer}{edgelayer}
		\draw  [style = thick](0.center) to (2.center);
		\draw  [style = thick](2.center) to (1.center);
		\draw  [style = thick](1.center) to (0.center);
		\draw  [style = thick](3.center) to (4.center);
		\draw  [style = thick](6.center) to (8.center);
		\draw  [style = thick](16.center) to (11.center);
		\draw  [style = thick](11.center) to (13.center);
		\draw  [style = thick](13.center) to (16.center);
		\draw  [style = thick](12.center) to (15.center);
		\draw  [style = thick](17.center) to (9.center);
		\draw  [style = thick](19.center) to (20.center);
		\draw  [style = thick](21.center) to (26.center);
		\draw  [style = thick](26.center) to (24.center);
		\draw  [style = thick](24.center) to (21.center);
		\draw  [style = thick](23.center) to (25.center);
	\end{pgfonlayer}
\end{tikzpicture}}
\endpgfgraphicnamed}  
\end{center} 

The $\epsilon$ and $\eta$ maps are depicted by cups and caps, and yanking by straightening of curves. We have one of the following tuple of diagrams for each $\epsilon$ and $\eta$:
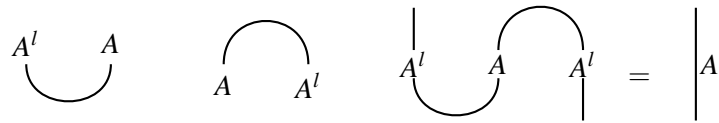
 \begin{center}
  {%
\beginpgfgraphicnamed{compact-cap-cup}
\begin{tikzpicture}[scale=0.75]
	\begin{pgfonlayer}{nodelayer}
		\node [style=none] (0) at (-5, 0) {};
		\node [style=none] (1) at (-2, 0) {};
		\node [style=none] (2) at (-5, 0.75) {$A^l$};
		\node [style=none] (3) at (2, -0.75) {$A$};
		\node [style=none] (4) at (5, -0.75) {$A^l$};
		\node [style=none] (5) at (2, 0) {};
		\node [style=none] (6) at (5, 0) {};
		\node [style=none] (7) at (-2, 0.75) {$A$};
	\end{pgfonlayer}
	\begin{pgfonlayer}{edgelayer}
		\draw [thick, bend right=90, looseness=1.50] (0.center) to (1.center);
		\draw [thick, bend left=90, looseness=1.75] (5.center) to (6.center);
	\end{pgfonlayer}
\end{tikzpicture}}
\endpgfgraphicnamed}
  \qquad
    {%
\beginpgfgraphicnamed{compact-yank}
\begin{tikzpicture}[scale=0.75]
	\begin{pgfonlayer}{nodelayer}
		\node [style=none] (0) at (-5, 0) {};
		\node [style=none] (1) at (-2, 0) {};
		\node [style=none] (2) at (-5, 0.5) {$A^l$};
		\node [style=none] (3) at (-2, 0.5) {$A$};
		\node [style=none] (4) at (1, 0.5) {$A^l$};
		\node [style=none] (5) at (-2, 1) {};
		\node [style=none] (6) at (1, 1) {};
		\node [style=none] (7) at (3, 0) {$=$};
		\node [style=none] (8) at (5, 2.5) {};
		\node [style=none] (9) at (5, -1.5) {};
		\node [style=none] (10) at (5.5, 0.5) {$A$};
		\node [style=none] (11) at (-5, 1) {};
		\node [style=none] (12) at (-5, 2.5) {};
		\node [style=none] (13) at (1, 0) {};
		\node [style=none] (14) at (1, -1.5) {};
	\end{pgfonlayer}
	\begin{pgfonlayer}{edgelayer}
		\draw [thick, bend right=90, looseness=1.50] (0.center) to (1.center);
		\draw [thick, bend left=90, looseness=1.75] (5.center) to (6.center);
		\draw [style=thick] (8.center) to (9.center);
		\draw [style=thick] (12.center) to (11.center);
		\draw [style=thick] (13.center) to (14.center);
	\end{pgfonlayer}
\end{tikzpicture}}
\endpgfgraphicnamed}
\end{center}

The diagrams for the bialgebraic  monoid and  comonoid 
morphisms and their interaction (the bialgebra law Q3) are as follows:

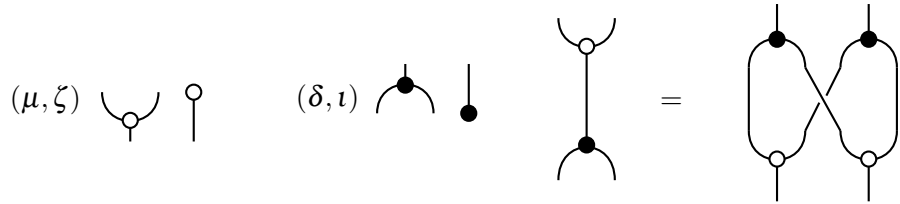
\begin{center}
{%
\beginpgfgraphicnamed{comp-alg-coalg}
\begin{tikzpicture}[scale=0.75]
	\begin{pgfonlayer}{nodelayer}
		\node [style=none] (0) at (-3.5, 2.25) {};
		\node [draw, thick, style=none, minimum size=0.2 cm, circle, fill=white] (1) at (-4.5, 1.25) {};
		\node [style=none] (2) at (-5.5, 2.25) {};
		\node [style=none] (3) at (-4.5, 0.5) {};
		\node [style=none] (4) at (4.25, 1.5) {};
		\node [style=none] (5) at (6.25, 1.5) {};
		\node [style=none] (6) at (4.25, 1.5) {};
		\node [draw, thick, style=none, minimum size=0.2 cm, circle, fill=black] (7) at (5.25, 2.5) {};
		\node [style=none] (8) at (5.25, 3.25) {};
		\node [style=none] (9) at (6.25, 1.5) {};
		\node [style=none] (10) at (-7.5, 2) {$(\mu,\zeta)$};
		\node [draw, thick, style=none, minimum size=0.2 cm, circle, fill=white] (11) at (-2.25, 2.25) {};
		\node [style=none] (12) at (-2.25, 0.5) {};
		\node [style=none] (13) at (2.5, 2) {$(\delta,\iota)$};
		\node [draw, thick, style=none, minimum size=0.2 cm, circle, fill=black] (14) at (7.5, 1.5) {};
		\node [style=none] (15) at (7.5, 3.25) {};
	\end{pgfonlayer}
	\begin{pgfonlayer}{edgelayer}
		\draw [style=thick] (1.center) to (3.center);
		\draw [style=thick, bend left=90, looseness=1.75] (6.center) to (5.center);
		\draw [style=thick] (8.center) to (7.center);
		\draw [thick, bend right=90, looseness=1.75] (2.center) to (0.center);
		\draw [style=thick] (11.center) to (12.center);
		\draw [style=thick] (15.center) to (14.center);
	\end{pgfonlayer}
\end{tikzpicture}}
\endpgfgraphicnamed} \qquad {%
\beginpgfgraphicnamed{bialg-equation}
\begin{tikzpicture}[scale=0.75]
	\begin{pgfonlayer}{nodelayer}
		\node [style=none] (0) at (-3, 0.5) {$=$};
		\node [style=none] (1) at (-5, 3.5) {};
		\node [style=none] (2) at (-7, 3.5) {};
		\node [draw, thick, style=none, minimum size=0.2 cm, circle, fill=white] (3) at (-6, 2.5) {};
		\node [style=none] (4) at (-7, -2.25) {};
		\node [style=none] (5) at (-5, -2.25) {};
		\node [draw, thick, style=none, minimum size=0.2 cm, circle, fill=black] (6) at (-6, -1) {};
		\node [style=none] (7) at (-0.25, -0.5) {};
		\node [style=none] (8) at (-0.25, 1.75) {};
		\node [style=none] (9) at (1.75, 1.75) {};
		\node [style=none] (10) at (-0.25, 1.75) {};
		\node [draw, thick, style=none, minimum size=0.2 cm, circle, fill=black] (11) at (0.75, 2.75) {};
		\node [style=none] (12) at (0.75, 4) {};
		\node [style=none] (13) at (1.75, 1.75) {};
		\node [style=none] (14) at (5, 1.75) {};
		\node [draw, thick, style=none, minimum size=0.2 cm, circle, fill=black] (15) at (4, 2.75) {};
		\node [style=none] (16) at (3, 1.75) {};
		\node [style=none] (17) at (4, 4) {};
		\node [style=none] (18) at (5, 1.75) {};
		\node [style=none] (19) at (3, 1.75) {};
		\node [draw, thick, style=none, minimum size=0.2 cm, circle, fill=white] (20) at (0.75, -1.5) {};
		\node [style=none] (21) at (1.75, -0.5) {};
		\node [style=none] (22) at (-0.25, -0.5) {};
		\node [draw, thick, style=none, minimum size=0.2 cm, circle, fill=white] (23) at (4, -1.5) {};
		\node [style=none] (24) at (5, -0.5) {};
		\node [style=none] (25) at (3, -0.5) {};
		\node [style=none] (26) at (0.75, -1.5) {};
		\node [style=none] (27) at (4, -1.5) {};
		\node [style=none] (28) at (5, -0.5) {};
		\node [style=none] (29) at (5, 1.75) {};
		\node [style=none] (30) at (1.75, 1.75) {};
		\node [style=none] (31) at (3, 1.75) {};
		\node [style=none] (32) at (2.5, 0.75) {};
		\node [style=none] (33) at (0.75, -3) {};
		\node [style=none] (34) at (4, -3) {};
		\node [style=none] (35) at (2.25, 0.5) {};
	\end{pgfonlayer}
	\begin{pgfonlayer}{edgelayer}
		\draw [thick, bend left=90, looseness=1.75] (1.center) to (2.center);
		\draw [style=thick, bend left=90, looseness=2.00] (4.center) to (5.center);
		\draw [style=thick, in=90, out=270] (3.center) to (6.center);
		\draw [style=thick] (7.center) to (8.center);
		\draw [style=thick, in=90, out=90, looseness=1.75] (10.center) to (9.center);
		\draw [style=thick] (12.center) to (11.center);
		\draw [style=thick, in=90, out=90, looseness=1.75] (16.center) to (14.center);
		\draw [style=thick] (17.center) to (15.center);
		\draw [thick, bend left=90, looseness=1.75] (21.center) to (22.center);
		\draw [thick, bend left=90, looseness=1.75] (24.center) to (25.center);
		\draw [style=thick] (29.center) to (28.center);
		\draw [style=thick] (13.center) to (25.center);
		\draw [style=thick] (31.center) to (32.center);
		\draw [style=thick] (26.center) to (33.center);
		\draw [style=thick] (27.center) to (34.center);
		\draw [style=thick] (35.center) to (21.center);
	\end{pgfonlayer}
\end{tikzpicture}}
\endpgfgraphicnamed}
\end{center}

\section{Generalised Quantifiers in Natural Language}
A generalised quantifier  $q$ on  a universe $U$  is  the image of a  function of the following form:
\[
q \colon {\cal P}(U) \to {\cal P}{\cal P}(U)
\]
We sometimes abuse the terminology and   call the function itself a quantifier.   The $q$  function can equivalently be represented  by a relation  over ${\cal P}(U)$, that is by a subset of ${\cal P}(U) \times {\cal P}(U)$. We use a  category theoretical notation and denote this  relation by  a barred line as follows: ${\cal P}(U) \relto {\cal P}(U)$. 

  For $A$ and $X$  subsets of $U$, examples of this relation for first order quantifiers are as follows:

\[A \stackrel{\text{some}}{\relto}  X  \quad   \iff \quad X \cap A \neq \emptyset \qquad \qquad
A \stackrel{\text{all}}{\relto}  X  \quad   \iff \quad A \subseteq X\\
\]
Quantifiers that go beyond first order,  for example  `few, most, several, many' are also definable. For $\alpha$ such a quantifier, we have:
\[
A \stackrel{{\alpha}}{\relto}  X  \quad   \iff \quad |A \cap X| = \alpha \ \mbox{elements of} \ A
\]
where the definition of `$\alpha$ elements of a set' depends on one's  underlying model.  The following property is of importance when using such quantifiers in natural language. A quantifier  $q$ is said to be \emph{conservative} whenever   $A \stackrel{q}{\relto} X$ iff  $A \stackrel{q}{\relto}  X \cap A$. If a quantifier $q$ is conservative, it is   said that  $q$ lives on $A$. 

Syntactically, quantified phrases of natural language are generated via the following context free rules, where NP is a noun phrase, N is a common noun, and D is a determiner:
\[
\text{NP} \to \text{D}  \ \text{N} \qquad \text{N} \to \mbox{cat, dog, men}, \cdots \qquad
\text{D} \to \mbox{a, some, all, no, most, few}, \cdots
\]

In \cite{BarwiseCooper81}, Barwise and Cooper  take the semantics of natural language to be a pair $(U, \semantics{\ })$, where $U$ is a universe of reference  and  $\semantics{\ }$ is a map from the vocabulary of the language to subsets or sets of subsets of $U$ or products thereof. This assignment depends on the types of the words.  The semantics of a verb phrase $vp$  is a unary relation over $U$, that is $\semantics{vp} \subseteq U$, that of a verb is a binary relation, that is $\semantics{v} \subseteq U \times U$. 
The semantics of a  determiner $d$   is   a map of the form $\semantics{d} \colon {\cal P}(U) \to  {\cal P}{\cal P}(U)$.  Given $\semantics{n}$ as the   semantics of a noun $n$, that is $\semantics{n} \subseteq U$, the  semantics of a quantified noun  $\semantics{d \ n}$  is defined to be $\semantics{d}(\semantics{n})$ and   is  a set of subsets, that is $\semantics {d \ n} \subseteq {\cal P}{\cal P}(U)$. Here, $d$ is a conservative generalised quantifier. To be conservative for $d$  means that the following holds: 

\[
\semantics{n} \stackrel{{\tiny d}}{\relto} X \iff  \semantics{n} \stackrel{d}{\relto} X \cap \semantics{n}
\]

The semantics of    phrases  and sentences are defined by  induction  over their   generation structure.  The semantics of a sentence $s$  is referred to by `true'  whenever $\semantics{s}  \neq \emptyset$.

\section{Bialgebraic Treatment of Generalised Quantifiers}
In previous work \cite{HedgesSadr},  we presented a compositional way of representing quantified phrases and sentences of natural language as morphisms of a self adjoint compact closed category $\cal C$ that has two designated objects $W$ and $S$,  where $W$ has a bialgebra structure over it.  In order to make the syntax-semantics interface work, we started with a context free grammar, turned it into a pregroup grammar (via a known translation \cite{Buszkowski-prg}), and defined a strongly monoidal functor  $\overline{\semantics{ \ }}$ between this pregroup grammar and  $\cal C$ as described above. We used the tuple $({\cal C}, {W,S}, \overline{\semantics{\ }})$  to represent this interface.  In the current paper, we shall skip  the syntactic details and only review  the semantics. 

A determiner $d$ is interpreted using  a morphism of the form $\ovl{\semantics{d}} \colon W \to W$.  The meaning of a determiner-noun phrase `$d \ n$', becomes 
$\overline{\semantics{d \ n}} = \ovl{\semantics{d}} \circ \ovl{\semantics{n}}$. In order to express conservativity of quantifiers, we need an intersection operator, so we use the bialgebraic internal monoid  map $\mu$. One needs to  perform  another action,  other than intersection,  on  the input of  the $d$ map. We  first need to apply the $\ovl{\semantics{d}}$ morphism to this input,  and only then take  its intersection with the result. So we need to first copy the input.  For this,   we use the bialgebraic  comonoid map $\delta$.   With these considerations in mind, we defined a  conservative quantifier to  be the following morphism:
{\small
\[
\ovl{\semantics{d}}   = (1_W \otimes \epsilon_W)
\circ (1_W \otimes \mu_W \otimes \epsilon_W \otimes 1_W)
\circ (1_W \otimes \overline{\semantics{d}} \otimes \delta_W \otimes 1_{W\otimes W})
 \circ (1_W \otimes \eta_W \otimes 1_{W \otimes W})
 \circ (\eta_W \otimes 1_W)
\]}
which is depicted as follows: 
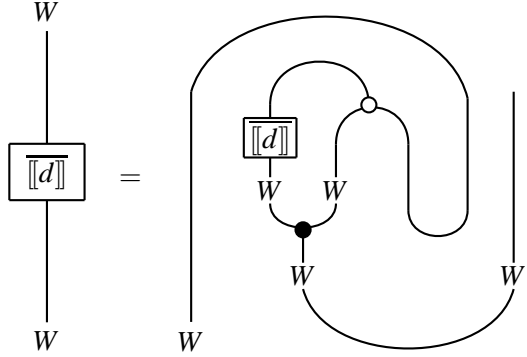
\begin{center}
\begin{minipage}{1cm}{%
\beginpgfgraphicnamed{Det-N-simple}
\begin{tikzpicture}
	\begin{pgfonlayer}{nodelayer}
		\node [style=none] (0) at (-1, -1) {};
		\node [style=none] (1) at (0, -1) {};
		\node [style=none] (2) at (0, 0.5) {};
		\node [style=none] (3) at (1, 0.5) {};
		\node [style=none] (4) at (1, -1) {};
		\node [style=none] (5) at (0, -0.25) {$\overline{\semantics{d}}$};
		\node [style=none] (6) at (0, 3.5) {};
		\node [style=none] (7) at (0, -4.25) {};
		\node [style=none] (8) at (0, -4.75) {$W$};
		\node [style=none] (9) at (0, 4) {$W$};
		\node [style=none] (10) at (-1, 0.5) {};
	\end{pgfonlayer}
	\begin{pgfonlayer}{edgelayer}
		\draw [style=thick] (10.center) to (3.center);
		\draw [style=thick] (3.center) to (4.center);
		\draw [style=thick] (4.center) to (0.center);
		\draw [style=thick] (0.center) to (10.center);
		\draw [style=thick] (1.center) to (7.center);
		\draw [style=thick] (6.center) to (2.center);
	\end{pgfonlayer}
\end{tikzpicture}}
\endpgfgraphicnamed} \end{minipage}
\quad $=$ \quad 
\begin{minipage}{5cm}{%
\beginpgfgraphicnamed{Det-N}
\begin{tikzpicture}[scale=0.7]
	\begin{pgfonlayer}{nodelayer}
		\node [draw, thick, style=none, minimum size=0.2 cm, circle, fill=black] (0) at (-4.75, -1.25) {};
		\node [style=none] (1) at (-6, 3.5) {};
		\node [style=none] (2) at (-2.25, 3.5) {};
		\node [style=none] (3) at (-7, 3) {};
		\node [style=none] (4) at (-5, 3) {};
		\node [style=none] (5) at (-5, 1.5) {};
		\node [style=none] (6) at (-7, 1.5) {};
		\node [style=none] (7) at (-6, 1.5) {};
		\node [style=none] (8) at (-4.75, -2.5) {};
		\node [style=none] (9) at (-6, 2.25) {$\overline{\semantics{d}}$};
		\node [style=none] (10) at (-4.75, -3) {$W$};
		\node [style=none] (11) at (-4.75, -1.25) {};
		\node [style=none] (12) at (-6, 0.75) {};
		\node [style=none] (13) at (-6, 3) {};
		\node [style=none] (14) at (-6, 0.25) {$W$};
		\node [style=none] (15) at (-6, -0.25) {};
		\node [style=none] (16) at (-3.5, -0.25) {};
		\node [style=none] (17) at (-3.5, 0.25) {$W$};
		\node [style=none] (18) at (-9, 4) {};
		\node [style=none] (19) at (-9, -4.75) {};
		\node [style=none] (20) at (-9, -5.5) {$W$};
		\node [style=none] (21) at (3.25, -3) {$W$};
		\node [style=none] (22) at (3.25, 6.75) {};
		\node [style=none] (23) at (3.25, -2.5) {};
		\node [style=none] (24) at (-4.75, -3.5) {};
		\node [style=none] (25) at (3.25, -3.5) {};
		\node [style=none] (27) at (-3.5, 2) {};
		\node [style=none] (28) at (-0.75, -0.5) {};
		\node [draw, thick, style=none, minimum size=0.2 cm, circle, fill=white] (29) at (-2.25, 3.5) {};
		\node [style=none] (30) at (-3.5, 2) {};
		\node [style=none] (31) at (-3.5, 0.75) {};
		\node [style=none] (32) at (1.5, -0.5) {};
		\node [style=none] (33) at (1.5, 3.75) {};
		\node [style=none] (34) at (-0.75, -0.5) {};
		\node [style=none] (35) at (-0.75, 2) {};
		\node [style=none] (36) at (3.25, 4) {};
	\end{pgfonlayer}
	\begin{pgfonlayer}{edgelayer}
		\draw [style=thick, in=90, out=90, looseness=1.50] (1.center) to (2.center);
		\draw [style=thick] (3.center) to (4.center);
		\draw [style=thick] (4.center) to (5.center);
		\draw [style=thick] (5.center) to (6.center);
		\draw [style=thick] (6.center) to (3.center);
		\draw [style=thick] (11.center) to (8.center);
		\draw [style=thick] (7.center) to (12.center);
		\draw [style=thick, bend right=90, looseness=1.25] (15.center) to (16.center);
		\draw [style=thick, bend right=75] (24.center) to (25.center);
		\draw [style=thick] (30.center) to (31.center);
		\draw [style=thick] (18.center) to (19.center);
		\draw [style=thick] (1.center) to (13.center);
		\draw [style=thick, bend right=90, looseness=1.50] (28.center) to (32.center);
		\draw [style=thick] (33.center) to (32.center);
		\draw [style=thick, bend left=75] (18.center) to (33.center);
		\draw [style=thick] (35.center) to (34.center);
		\draw [style=thick, bend left=90, looseness=1.75] (30.center) to (35.center);
		\draw [style=thick] (36.center) to (23.center);
	\end{pgfonlayer}
\end{tikzpicture}}
\endpgfgraphicnamed} \end{minipage} 
\end{center}

\noindent
Taking the above into account, we obtain the following for the meaning of  `d n':

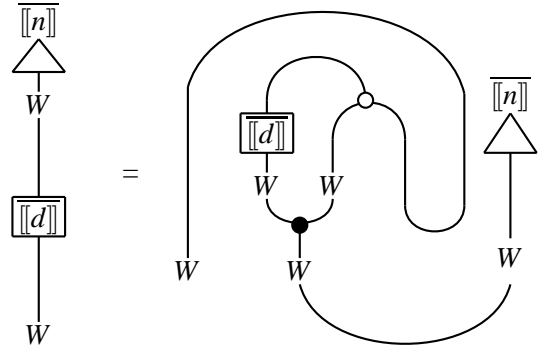
\begin{center}
\begin{minipage}{1cm}
{%
\beginpgfgraphicnamed{Det-N-triangle}
\begin{tikzpicture}[scale=0.7]
	\begin{pgfonlayer}{nodelayer}
		\node [style=none] (0) at (-4.25, -3) {};
		\node [style=none] (1) at (-3.25, -3) {};
		\node [style=none] (2) at (-3.25, -1.5) {};
		\node [style=none] (3) at (-2.25, -1.5) {};
		\node [style=none] (4) at (-2.25, -3) {};
		\node [style=none] (5) at (-3.25, -2.25) {$\overline{\semantics{d}}$};
		\node [style=none] (6) at (-3.25, 1.5) {};
		\node [style=none] (7) at (-3.25, -6.25) {};
		\node [style=none] (8) at (-3.25, -6.75) {$W$};
		\node [style=none] (9) at (-3.25, 2) {$W$};
		\node [style=none] (10) at (-4.25, -1.5) {};
		\node [style=none] (11) at (-3.25, 2.5) {};
		\node [style=none] (12) at (-3.25, 4.5) {};
		\node [style=none] (13) at (-2.25, 3.25) {};
		\node [style=none] (14) at (-4.25, 3.25) {};
		\node [style=none] (15) at (-3.25, 3.25) {};
		\node [style=none] (16) at (-3.25, 5.25) {$\ovl{\semantics{n}}$};
	\end{pgfonlayer}
	\begin{pgfonlayer}{edgelayer}
		\draw [style=thick] (10.center) to (3.center);
		\draw [style=thick] (3.center) to (4.center);
		\draw [style=thick] (4.center) to (0.center);
		\draw [style=thick] (0.center) to (10.center);
		\draw [style=thick] (1.center) to (7.center);
		\draw [style=thick] (6.center) to (2.center);
		\draw [style=thick] (14.center) to (12.center);
		\draw [style=thick] (12.center) to (13.center);
		\draw [style=thick] (13.center) to (14.center);
		\draw [style=thick] (15.center) to (11.center);
	\end{pgfonlayer}
\end{tikzpicture}}
\endpgfgraphicnamed}
\end{minipage}  \quad = \quad 
\begin{minipage}{5cm}
{%
\beginpgfgraphicnamed{Q-noun}
\begin{tikzpicture}[scale=0.7]
	\begin{pgfonlayer}{nodelayer}
		\node [style=none] (0) at (6.25, 2.5) {};
		\node [style=none] (1) at (8.25, 2.5) {};
		\node [style=none] (2) at (7.25, 4) {};
		\node [style=none] (3) at (7.25, 4.75) {$\overline{\semantics{n}}$};
		\node [style=none] (4) at (7.25, 2.5) {};
		\node [style=none] (5) at (7.25, -0.75) {};
		\node [style=none] (6) at (7.25, -1.5) {$W$};
		\node [draw, thick, style=none, minimum size=0.2 cm, circle, fill=black] (7) at (-0.75, -0.25) {};
		\node [style=none] (8) at (-2, 1.25) {$W$};
		\node [style=none] (9) at (-0.75, -2.5) {};
		\node [style=none] (10) at (1.75, 4.5) {};
		\node [style=none] (11) at (0.5, 3) {};
		\node [style=none] (12) at (-3, 2.5) {};
		\node [style=none] (13) at (-0.75, -0.25) {};
		\node [style=none] (14) at (-3, 4) {};
		\node [style=none] (15) at (-2, 4.5) {};
		\node [style=none] (16) at (7.25, -2.5) {};
		\node [style=none] (17) at (-2, 3.25) {$\overline{\semantics{d}}$};
		\node [style=none] (18) at (-5, 5) {};
		\node [style=none] (19) at (-2, 4) {};
		\node [style=none] (20) at (-0.75, -2) {$W$};
		\node [style=none] (21) at (0.5, 1.75) {};
		\node [style=none] (22) at (-1, 2.5) {};
		\node [style=none] (23) at (0.5, 1.25) {$W$};
		\node [style=none] (24) at (-2, 2.5) {};
		\node [style=none] (25) at (5.5, 4.75) {};
		\node [style=none] (26) at (-2, 0.75) {};
		\node [style=none] (27) at (0.5, 3) {};
		\node [draw, thick, style=none, minimum size=0.2 cm, circle, fill=white] (28) at (1.75, 4.5) {};
		\node [style=none] (29) at (-0.75, -1.5) {};
		\node [style=none] (30) at (-1, 4) {};
		\node [style=none] (31) at (3.25, 0.5) {};
		\node [style=none] (32) at (5.5, 0.5) {};
		\node [style=none] (33) at (-5, -1.5) {};
		\node [style=none] (34) at (-2, 1.75) {};
		\node [style=none] (35) at (3.25, 3) {};
		\node [style=none] (36) at (0.5, 0.75) {};
		\node [style=none] (37) at (-5, -2) {$W$};
		\node [style=none] (38) at (3.25, 0.5) {};
	\end{pgfonlayer}
	\begin{pgfonlayer}{edgelayer}
		\draw  [style=thick] (0.center) to (1.center);
		\draw   [style=thick] (2.center) to (0.center);
		\draw  [style=thick]  (2.center) to (1.center);
		\draw  [style=thick]  (4.center) to (5.center);
		\draw [style=thick, in=90, out=90, looseness=1.50] (15.center) to (10.center);
		\draw [style=thick] (14.center) to (30.center);
		\draw [style=thick] (30.center) to (22.center);
		\draw [style=thick] (22.center) to (12.center);
		\draw [style=thick] (12.center) to (14.center);
		\draw [style=thick] (13.center) to (29.center);
		\draw [style=thick] (24.center) to (34.center);
		\draw [style=thick, bend right=90, looseness=1.25] (26.center) to (36.center);
		\draw [style=thick, bend right=75] (9.center) to (16.center);
		\draw [style=thick] (27.center) to (21.center);
		\draw [style=thick] (18.center) to (33.center);
		\draw [style=thick] (15.center) to (19.center);
		\draw [style=thick, bend right=90, looseness=1.50] (38.center) to (32.center);
		\draw [style=thick] (25.center) to (32.center);
		\draw [style=thick, bend left=75] (18.center) to (25.center);
		\draw [style=thick] (35.center) to (31.center);
		\draw [style=thick, bend left=90, looseness=1.75] (27.center) to (35.center);
	\end{pgfonlayer}
\end{tikzpicture}}
\endpgfgraphicnamed}
\end{minipage}
\end{center}

\noindent
Using the dual of the bialgebra defined previously, the above can further simplify.  or reasons of space, we do not give this simplification here and refer the reader to \cite{HedgesSadr}. 



So far, we have worked with an abstract compact closed categorical setting and within tuples of the form $({\cal C}, {W,S}, \overline{\semantics{\ }})$, as defined previously.   This abstract setting can instantiate to provide concrete models. For instance,  a  \emph{ relational  instantiation}    of the abstract  setting can be the tuple   $(\Rel, \cal P (U), \{\star\}, \overline{\semantics{\ }})$ with $ \cal P (U)$  and the singleton  set $\{\star\}$ as its two designated objects and where  ${\cal P}(U)$ has a bialgebra over it.  This is the relational instantiation we used in previous work to prove an equivalence between  the truth theoretic version of our semantics and that of Barwise and Cooper.  Herein, we first defined meaning of a sentence $s$  to be true  iff  $\star \ovl{\semantics{s}} \star$. We then proved  the following equivalence:
\begin{equation}
\label{thm}
\star \ovl{\semantics{s}} \star  \quad \text{iff} \quad \semantics{s} \neq \emptyset
\end{equation}
That is,  the meaning of a  quantified sentence is true in the relational instantiation of our abstract categorical setting  iff it is true in the generalised quantifier theory.  Since sets and relations embed into vector spaces and linear maps (sets as vector spaces spanned by their elements and relations as linear maps corresponding to their tables), one can immediately obtains the following embedding  of instantiations:
\[
(\Rel, \cal P (U), \{\star\}, \overline{\semantics{\ }}) \quad \leadsto \quad (\text{FdVect}, V_{{\cal P(U)}},V_{\{\star\}}, \ovl{\semantics{\ }})
\]
where FdVect here is the category of  vector spaces equipped with fixed orthogonal  bases indexed by finite sets from the universe $U$.  In FdVect, the vector version of equivalence (1) holds, for details please see \cite{HedgesSadr}. Since $V_{\{\star\}} \cong \mathbb{R}$, meanings of sentences in this model become real numbers, interpretable as degrees of truth.  In order to have vectors as meanings of sentences, we instantiated the model to tuples of the form $(\text{FdVect}, V_{{\cal P}(\Sigma)},Z, \ovl{\semantics{\text{\ }}})$, where $Z$ is a vector space wherein interpretations of  sentences live. Again, for details please see \cite{HedgesSadr}.

\section{Scope}
For the sake of explaining the  question of scope, consider a natural language sentence $d_1 n_1 v d_2 n_2$ with two determiners $d_1$ and $d_2$, a subject $n_1$, an object $n_2$,  and a  verb  $v$.   Suppose the semantics of this sentence is represented in a logical form, where  the words and their semantics are denoted by the same letter. So  subject 1 and its semantics are both denoted by $n_1$ and similarly for the rest of the words.  One faces two possibilities in the semantics regarding the scopes of the quantifications, as follows:
\[
(1)\ {d}_1 x \big({n}_1(x), {d}_2 y ({n}_2 (y), {v}(x,y))\big)\qquad
(2)\  {d}_2 y \big({n}_2(y), {d}_1 x ({n}_1 (x), {v}(x,y))\big)
\]
In    (1),  $d_1$ has a wide scope and $d_2$ a narrow scope, whereas in (2) $d_2$ has a wide scope and $d_1$ a narrow scope. As an example, consider the sentence `all men admire some cars', which can be interpreted by either of  the following two formulae:
\[
(1)\ \forall x \big(man(x), \exists y (car (y), admire(x,y))\big)\qquad
(2)\  \exists y \big(car(y), \forall x (man (x), admire(x,y))\big)
 \]
 Generalised quantifiers cannot in general be represented in first order logic and the above  are not meant to be formulae of first order logic. To see this, note that we have used commas and not the usual conjunction/implication connectives inside the brackets. 
 
Depending on the choice of quantifiers, the above two  options might or might not provide the same semantics for the sentence. For instance, they will  amount to the same meaning for the sentence ``some men admire some cars'', whereas for the sentence `` all men admire some cars'', or ``all men admire two cars'', these interpretations result in two different meanings. For option (2)  to be true, all men have to admire the same car  in the sentence ``all men admire some cars'' and  they have to admire the same two cars in the sentence ``all men admire two cars''.  Due to the presence of such ambiguities in natural language sentences, it is desirable that a semantic system can represent both of these interpretations.    
The question we address in what follows  is whether  our bialgebraic treatment of generalised quantifiers reflects this  ambiguity  and if so how.

When dealing with the question of scope, the  form of the predicates, i.e. whether they are sets or vectors, relations or linear maps,  is less important than the general pattern of the sentences containing them. This pattern is the same in any instantiation of the abstract categorical semantics. Having this mind,  we start our treatment by working within the relational instantiation and at the end provide an abstract categorical solution using the diagrammatic representation. 

In the relational instantiation,    the above two readings are obtained by computing the  results of the following  two interpretations:
\[
(1)\  \ovl{\semantics{d_1 \ n_1}} \big(\ovl{\semantics{v \ d_2 \ n_ 2}}\big)\qquad
(2)\  \ovl{\semantics{d_2\ n_2}} \big(\ovl{\semantics{d_1 \ n_1 \ v}}\big)
\]
Unfolding the above provides us with the following:
\begin{eqnarray*}
&(1)\ &{ \semantics{n_1}} \stackrel{{\semantics{d_1}}}{\relto} 
\{a \in {\semantics{n_1}} \mid {\semantics{d_2 n_2 v_a}}\},  \ \mbox{where for } \ {\semantics{d_2 n_2 v_a}}  \ \mbox{we have} \  {\semantics{n_2}} \stackrel{{\semantics{d_2}}} {\relto} \{b \in {\semantics{n_2}} \mid a \stackrel{\semantics{v}}{\relto}b\}\\
&(2)\ &{ \semantics{n_2}} \stackrel{{\semantics{d_2}}}{\relto} 
\{b \in {\semantics{n_2}} \mid {\semantics{d_1 n_1 v^{-1}_b}}\},  \ \mbox{where for} \ {\semantics{d_1 n_1 v^{-1}_b}}  \ \mbox{we have} \ {\semantics{n_1}} \stackrel{{\semantics{d_1}}} {\relto} \{a \in {\semantics{n_1}} \mid b \stackrel{{\semantics{v}}^{-1}}{\relto} a\}
\end{eqnarray*}

\noindent
In option (1), we first compute the $n_2$'s (i.e. $b \in \semantics{n_2}$),  that are in the $v$ relationship with an $n_1$ and check which one of these are in relationship $d_2$ with $n_2$, that is, for example when this set is in $d_2(n_2)$. From these,  we pick the elements whose $n_1$'s (i.e. $a \in \semantics{n_1})$   are in relationship $d_1$ with $n_1$. In option (2), we do the same but in the opposite order: first pick $n_1$'s that are in the $v$ relationship with an $n_2$ and check which one of them are related to $d_1 (n_1)$, from these, we pick the ones whose $n_2$ is related to $d_2 (n_2)$.

In the relational instantiation of the  categorical setting, we worked with powersets and had  our interpreted objects be of the same type: that is of type ${\cal P}(U)$. Hence, checking if a set is an element of another set and returning it as the result if this is the case, becomes equivalent to applying the  bialgebraic comonoid operation $\delta$ of type $ \colon {\cal P}(U) \to {\cal P}(U) \times {\cal P}(U)$. In the  abstract compact closed semantics, this is the comonoid map on the designated object $W$, that is $\delta \colon W \to W \times W$.  Taking these into account,   we obtain the following  two diagrammatic interpretations for the above two readings:

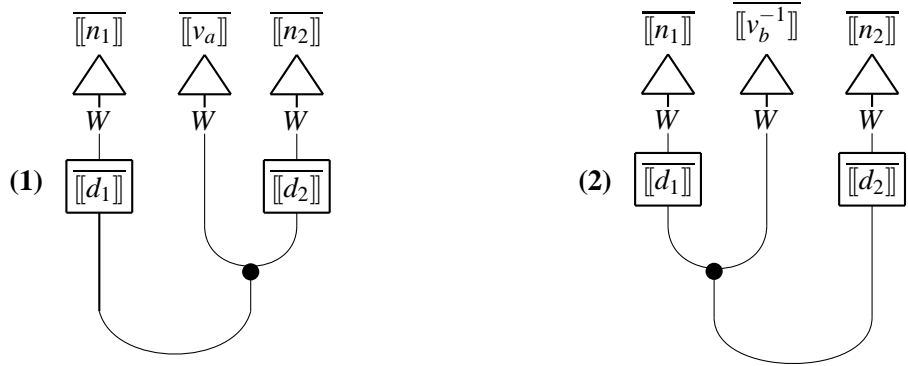
\begin{center}
{\bf (1)} \ {%
\beginpgfgraphicnamed{Reading1}
\begin{tikzpicture}[scale=0.7]
	\begin{pgfonlayer}{nodelayer}
		\node [style=none] (0) at (-5, 3.25) {};
		\node [style=none] (1) at (-5, 3.75) {};
		\node [style=none] (2) at (-5, 5.25) {};
		\node [style=none] (3) at (-4, 3.75) {};
		\node [style=none] (4) at (-5, 6.25) {$\overline{\semantics{n_1}}$};
		\node [style=none] (5) at (-6, 3.75) {};
		\node [style=none] (6) at (-5, 2.75) {$W$};
		\node [style=none] (7) at (-5, 2.25) {};
		\node [draw, thick, style=none, minimum size=0.2 cm, circle, fill=black] (8) at (0.75, -3) {};
		\node [style=none] (9) at (-1, 2.75) {$W$};
		\node [style=none] (10) at (-1, 3.25) {};
		\node [style=none] (11) at (-1, 3.75) {};
		\node [style=none] (12) at (-1, 6.25) {$\overline{\semantics{v_a}}$};
		\node [style=none] (13) at (-2, 3.75) {};
		\node [style=none] (14) at (0, 3.75) {};
		\node [style=none] (15) at (-1, 5.25) {};
		\node [style=none] (16) at (-1, 2.25) {};
		\node [style=none] (17) at (2.5, 2.75) {$W$};
		\node [style=none] (18) at (2.5, 3.25) {};
		\node [style=none] (19) at (2.5, 3.75) {};
		\node [style=none] (20) at (2.5, 6.25) {$\overline{\semantics{n_2}}$};
		\node [style=none] (21) at (1.5, 3.75) {};
		\node [style=none] (22) at (3.5, 3.75) {};
		\node [style=none] (23) at (2.5, 5.25) {};
		\node [style=none] (24) at (2.5, 2.25) {};
		\node [style=none] (25) at (2.5, -1.25) {};
		\node [style=none] (26) at (-1, -1.25) {};
		\node [style=none] (27) at (-5, -0.75) {};
		\node [style=none] (28) at (-3.75, 1.25) {};
		\node [style=none] (29) at (-5, -4.5) {};
		\node [style=none] (30) at (-5, 0.25) {$\overline{\semantics{d_1}}$};
		\node [style=none] (31) at (-6.25, 1.25) {};
		\node [style=none] (32) at (-5, 2.25) {};
		\node [style=none] (33) at (-5, 1.25) {};
		\node [style=none] (34) at (-6.25, -0.75) {};
		\node [style=none] (35) at (-3.75, -0.75) {};
		\node [style=none] (36) at (0.75, -4.5) {};
		\node [style=none] (37) at (2.5, -0.75) {};
		\node [style=none] (38) at (3.75, -0.75) {};
		\node [style=none] (39) at (3.75, 1.25) {};
		\node [style=none] (40) at (1.25, -0.75) {};
		\node [style=none] (41) at (1.25, 1.25) {};
		\node [style=none] (42) at (2.5, 1.25) {};
		\node [style=none] (43) at (2.5, 0.25) {$\overline{\semantics{d_2}}$};
	\end{pgfonlayer}
	\begin{pgfonlayer}{edgelayer}
		\draw [style=thick] (5.center) to (3.center);
		\draw [style=thick] (2.center) to (5.center);
		\draw [style=thick] (2.center) to (3.center);
		\draw [style=thick] (1.center) to (0.center);
		\draw [style=thick] (13.center) to (14.center);
		\draw [style=thick] (15.center) to (13.center);
		\draw [style=thick] (15.center) to (14.center);
		\draw [style=thick] (11.center) to (10.center);
		\draw [style=thick] (21.center) to (22.center);
		\draw [style=thick] (23.center) to (21.center);
		\draw [style=thick] (23.center) to (22.center);
		\draw [style=thick] (19.center) to (18.center);
		\draw (16.center) to (26.center);
		\draw [bend right=90, looseness=1.50] (26.center) to (25.center);
		\draw [style=thick] (31.center) to (28.center);
		\draw [style=thick] (28.center) to (35.center);
		\draw [style=thick] (35.center) to (34.center);
		\draw [style=thick] (34.center) to (31.center);
		\draw [style=thick] (27.center) to (29.center);
		\draw (32.center) to (33.center);
		\draw (8.center) to (36.center);
		\draw [bend right=75] (29.center) to (36.center);
		\draw [style=thick] (41.center) to (39.center);
		\draw [style=thick] (39.center) to (38.center);
		\draw [style=thick] (38.center) to (40.center);
		\draw [style=thick] (40.center) to (41.center);
		\draw (24.center) to (42.center);
		\draw (37.center) to (25.center);
	\end{pgfonlayer}
\end{tikzpicture}}
\endpgfgraphicnamed}
\hspace{3cm}
{\bf (2)} \ {%
\beginpgfgraphicnamed{Reading2}
\begin{tikzpicture}[scale=0.7]
	\begin{pgfonlayer}{nodelayer}
		\node [style=none] (0) at (2, 3.5) {};
		\node [style=none] (1) at (2, 4) {};
		\node [style=none] (2) at (2, 5.5) {};
		\node [style=none] (3) at (3, 4) {};
		\node [style=none] (4) at (2, 6.5) {$\overline{\semantics{n_2}}$};
		\node [style=none] (5) at (1, 4) {};
		\node [style=none] (6) at (2, 3) {$W$};
		\node [style=none] (7) at (2, 2.5) {};
		\node [draw, thick, style=none, minimum size=0.2 cm, circle, fill=black] (8) at (-4, -2.75) {};
		\node [style=none] (9) at (-2, 3) {$W$};
		\node [style=none] (10) at (-2, 3.5) {};
		\node [style=none] (11) at (-2, 4) {};
		\node [style=none] (12) at (-2, 6.75) {$\overline{\semantics{v_b^{-1}}}$};
		\node [style=none] (13) at (-3, 4) {};
		\node [style=none] (14) at (-1, 4) {};
		\node [style=none] (15) at (-2, 5.5) {};
		\node [style=none] (16) at (-2, 2.5) {};
		\node [style=none] (17) at (-5.75, 3) {$W$};
		\node [style=none] (18) at (-5.75, 3.5) {};
		\node [style=none] (19) at (-5.75, 4) {};
		\node [style=none] (20) at (-5.75, 6.5) {$\overline{\semantics{n_1}}$};
		\node [style=none] (21) at (-6.75, 4) {};
		\node [style=none] (22) at (-4.75, 4) {};
		\node [style=none] (23) at (-5.75, 5.5) {};
		\node [style=none] (24) at (-5.75, 2.5) {};
		\node [style=none] (25) at (-5.75, -1) {};
		\node [style=none] (26) at (-2, -1) {};
		\node [style=none] (27) at (2, -4.5) {};
		\node [style=none] (28) at (2, 2.5) {};
		\node [style=none] (29) at (-4, -4.5) {};
		\node [style=none] (30) at (-5.75, 1.75) {};
		\node [style=none] (31) at (-7, 1.75) {};
		\node [style=none] (32) at (-5.75, -0.25) {};
		\node [style=none] (33) at (-4.5, -0.25) {};
		\node [style=none] (34) at (-4.5, 1.75) {};
		\node [style=none] (35) at (-5.75, 0.75) {$\overline{\semantics{d_1}}$};
		\node [style=none] (36) at (-7, -0.25) {};
		\node [style=none] (37) at (2, 1.75) {};
		\node [style=none] (38) at (0.75, 1.75) {};
		\node [style=none] (39) at (2, -0.25) {};
		\node [style=none] (40) at (3.25, -0.25) {};
		\node [style=none] (41) at (3.25, 1.75) {};
		\node [style=none] (42) at (2, 0.75) {$\overline{\semantics{d_2}}$};
		\node [style=none] (43) at (0.75, -0.25) {};
	\end{pgfonlayer}
	\begin{pgfonlayer}{edgelayer}
		\draw [style=thick] (5.center) to (3.center);
		\draw [style=thick] (2.center) to (5.center);
		\draw [style=thick] (2.center) to (3.center);
		\draw [style=thick] (1.center) to (0.center);
		\draw [style=thick] (13.center) to (14.center);
		\draw [style=thick] (15.center) to (13.center);
		\draw [style=thick] (15.center) to (14.center);
		\draw [style=thick] (11.center) to (10.center);
		\draw [style=thick] (21.center) to (22.center);
		\draw [style=thick] (23.center) to (21.center);
		\draw [style=thick] (23.center) to (22.center);
		\draw [style=thick] (19.center) to (18.center);
		\draw (16.center) to (26.center);
		\draw (8.center) to (29.center);
		\draw [bend right=90, looseness=1.50] (25.center) to (26.center);
		\draw [bend right=90] (29.center) to (27.center);
		\draw [style=thick] (31.center) to (34.center);
		\draw [style=thick] (34.center) to (33.center);
		\draw [style=thick] (33.center) to (36.center);
		\draw [style=thick] (36.center) to (31.center);
		\draw [style=thick] (38.center) to (41.center);
		\draw [style=thick] (41.center) to (40.center);
		\draw [style=thick] (40.center) to (43.center);
		\draw [style=thick] (43.center) to (38.center);
		\draw (24.center) to (30.center);
		\draw (32.center) to (25.center);
		\draw (28.center) to (37.center);
		\draw (27.center) to (39.center);
	\end{pgfonlayer}
\end{tikzpicture}}
\endpgfgraphicnamed}
\end{center}

In the relational instantiation, we defined $\semantics{z_x}$, for $z$ either $v$ or $v^{-1}$ and $x \in X$,  to be the set $\{y \in Y \mid x \stackrel{\semantics{z}}{\relto} y\}$. The  abstract compact closed form of  this set is the morphism  $I \stackrel{\overline{\semantics{v_x}}}{\to}Y$, derivable  from the morphism  $I \stackrel{\overline{\semantics{v}}}{\to} X \otimes Y$. 
By conservativity, computing the truth of $\semantics{d \ n \ v_x}$ is the same as checking whether $\semantics{n} \cap \semantics{v_x} \in \semantics{d}(\semantics{n})$.  In the relational instantiation, we used the bialgebraic monoid map $\mu \colon {\cal P}(U)  \times   {\cal P}(U) \to {\cal P}(U) $ to model  intersection. In the abstract categorical setting, this becomes the monoid map on $W$, and the above diagrams  are unfolded as follows:

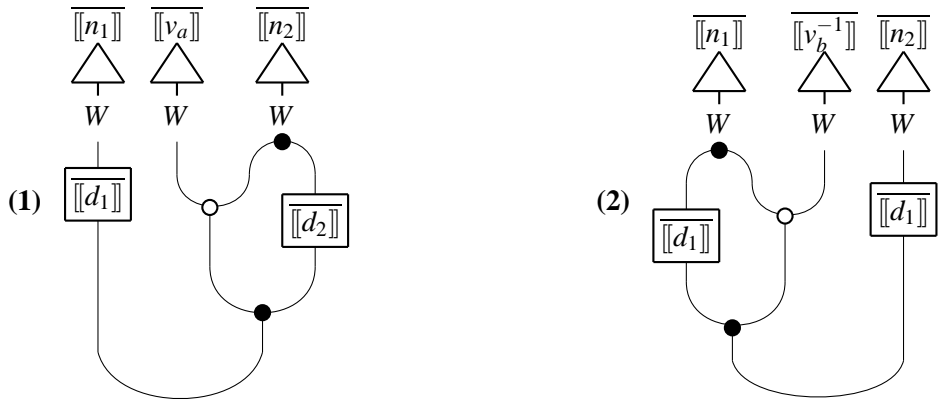
\begin{center}
{\bf (1)} \ {%
\beginpgfgraphicnamed{Reading1-opened}
\begin{tikzpicture}[scale=0.7]
	\begin{pgfonlayer}{nodelayer}
		\node [style=none] (0) at (-4, 4.25) {};
		\node [style=none] (1) at (-4, 4.75) {};
		\node [style=none] (2) at (-4, 6.25) {};
		\node [style=none] (3) at (-3, 4.75) {};
		\node [style=none] (4) at (-4, 7) {$\overline{\semantics{n_1}}$};
		\node [style=none] (5) at (-5, 4.75) {};
		\node [style=none] (6) at (-4, 3.5) {$W$};
		\node [style=none] (7) at (-4, 2.5) {};
		\node [draw, thick, style=none, minimum size=0.2 cm, circle, fill=black] (8) at (2.25, -4) {};
		\node [style=none] (9) at (-1, 3.5) {$W$};
		\node [style=none] (10) at (-1, 4.25) {};
		\node [style=none] (11) at (-1, 4.75) {};
		\node [style=none] (12) at (-1, 7) {$\overline{\semantics{v_a}}$};
		\node [style=none] (13) at (-2, 4.75) {};
		\node [style=none] (14) at (0, 4.75) {};
		\node [style=none] (15) at (-1, 6.25) {};
		\node [style=none] (16) at (0.25, 0) {};
		\node [style=none] (17) at (3, 3.5) {$W$};
		\node [style=none] (18) at (3, 4.25) {};
		\node [style=none] (19) at (3, 4.75) {};
		\node [style=none] (20) at (3, 7) {$\overline{\semantics{n_2}}$};
		\node [style=none] (21) at (2, 4.75) {};
		\node [style=none] (22) at (4, 4.75) {};
		\node [style=none] (23) at (3, 6.25) {};
		\node [style=none] (24) at (4.25, 1.25) {};
		\node [style=none] (25) at (4.25, -2.25) {};
		\node [style=none] (26) at (0.25, -2.25) {};
		\node [style=none] (27) at (-4, 2.5) {};
		\node [style=none] (28) at (2.25, -5.5) {};
		\node [draw, thick, style=none, minimum size=0.2 cm, circle, fill=white] (29) at (0.25, 0) {};
		\node [style=none] (30) at (1.75, 1.25) {};
		\node [style=none] (31) at (4.25, 1.25) {};
		\node [draw, thick, style=none, minimum size=0.2 cm, circle, fill=black] (32) at (3, 2.5) {};
		\node [style=none] (33) at (-1, 2.5) {};
		\node [style=none] (34) at (-1, 1.25) {};
		\node [style=none] (35) at (-4, -5.5) {};
		\node [style=none] (36) at (-4, 0.5) {$\overline{\semantics{d_1}}$};
		\node [style=none] (37) at (-5.25, 1.5) {};
		\node [style=none] (38) at (-4, 1.5) {};
		\node [style=none] (39) at (-4, -0.5) {};
		\node [style=none] (40) at (-2.75, 1.5) {};
		\node [style=none] (41) at (-2.75, -0.5) {};
		\node [style=none] (42) at (-5.25, -0.5) {};
		\node [style=none] (43) at (4.25, -0.5) {$\overline{\semantics{d_2}}$};
		\node [style=none] (44) at (3, 0.5) {};
		\node [style=none] (45) at (4.25, 0.5) {};
		\node [style=none] (46) at (4.25, -1.5) {};
		\node [style=none] (47) at (5.5, 0.5) {};
		\node [style=none] (48) at (5.5, -1.5) {};
		\node [style=none] (49) at (3, -1.5) {};
	\end{pgfonlayer}
	\begin{pgfonlayer}{edgelayer}
		\draw [style=thick] (5.center) to (3.center);
		\draw [style=thick] (2.center) to (5.center);
		\draw [style=thick] (2.center) to (3.center);
		\draw [style=thick] (1.center) to (0.center);
		\draw [style=thick] (13.center) to (14.center);
		\draw [style=thick] (15.center) to (13.center);
		\draw [style=thick] (15.center) to (14.center);
		\draw [style=thick] (11.center) to (10.center);
		\draw [style=thick] (21.center) to (22.center);
		\draw [style=thick] (23.center) to (21.center);
		\draw [style=thick] (23.center) to (22.center);
		\draw [style=thick] (19.center) to (18.center);
		\draw (16.center) to (26.center);
		\draw [bend right=90, looseness=1.50] (26.center) to (25.center);
		\draw (8.center) to (28.center);
		\draw [bend left=90, looseness=1.75] (30.center) to (31.center);
		\draw (33.center) to (34.center);
		\draw [bend right=90, looseness=1.50] (34.center) to (30.center);
		\draw [bend right=75] (35.center) to (28.center);
		\draw [style=thick] (37.center) to (40.center);
		\draw [style=thick] (40.center) to (41.center);
		\draw [style=thick] (41.center) to (42.center);
		\draw [style=thick] (42.center) to (37.center);
		\draw [style=thick] (44.center) to (47.center);
		\draw [style=thick] (47.center) to (48.center);
		\draw [style=thick] (48.center) to (49.center);
		\draw [style=thick] (49.center) to (44.center);
		\draw (27.center) to (38.center);
		\draw (39.center) to (35.center);
		\draw (31.center) to (45.center);
		\draw (46.center) to (25.center);
	\end{pgfonlayer}
\end{tikzpicture}}
\endpgfgraphicnamed}
\hspace{3cm}
{\bf (2)} \ {%
\beginpgfgraphicnamed{Reading2-opened}
\begin{tikzpicture}[scale=0.7]
	\begin{pgfonlayer}{nodelayer}
		\node [style=none] (0) at (-4, 4.25) {};
		\node [style=none] (1) at (-4, 4.75) {};
		\node [style=none] (2) at (-4, 6.25) {};
		\node [style=none] (3) at (-3, 4.75) {};
		\node [style=none] (4) at (-4, 7) {$\overline{\semantics{n_1}}$};
		\node [style=none] (5) at (-5, 4.75) {};
		\node [style=none] (6) at (-4, 3.5) {$W$};
		\node [style=none] (7) at (-5.25, 1.25) {};
		\node [draw, thick, style=none, minimum size=0.2 cm, circle, fill=black] (8) at (-3.5, -4.25) {};
		\node [style=none] (9) at (0, 3.5) {$W$};
		\node [style=none] (10) at (0, 4.25) {};
		\node [style=none] (11) at (0, 4.75) {};
		\node [style=none] (12) at (0, 7) {$\overline{\semantics{v_b^{-1}}}$};
		\node [style=none] (13) at (-1, 4.75) {};
		\node [style=none] (14) at (1, 4.75) {};
		\node [style=none] (15) at (0, 6.25) {};
		\node [style=none] (16) at (-1.5, 0) {};
		\node [style=none] (17) at (3, 3.5) {$W$};
		\node [style=none] (18) at (3, 4.25) {};
		\node [style=none] (19) at (3, 4.75) {};
		\node [style=none] (20) at (3, 7) {$\overline{\semantics{n_2}}$};
		\node [style=none] (21) at (2, 4.75) {};
		\node [style=none] (22) at (4, 4.75) {};
		\node [style=none] (23) at (3, 6.25) {};
		\node [style=none] (24) at (3, 2.5) {};
		\node [style=none] (25) at (3, -5.5) {};
		\node [style=none] (26) at (-1.5, -2.5) {};
		\node [style=none] (27) at (-5.25, 1.25) {};
		\node [style=none] (28) at (-3.5, -5.5) {};
		\node [draw, thick, style=none, minimum size=0.2 cm, circle, fill=white] (29) at (-1.5, 0) {};
		\node [style=none] (30) at (0, 1.25) {};
		\node [style=none] (31) at (3, 2.5) {};
		\node [style=none] (32) at (-2.75, 1.25) {};
		\node [style=none] (33) at (-5.25, -2.5) {};
		\node [style=none] (34) at (-5.25, 1.25) {};
		\node [draw, thick, style=none, minimum size=0.2 cm, circle, fill=black] (35) at (-4, 2.5) {};
		\node [style=none] (36) at (-2.75, 1.25) {};
		\node [style=none] (37) at (-2.75, 1.25) {};
		\node [style=none] (38) at (0, 2.5) {};
		\node [style=none] (39) at (-5.25, -0.75) {$\overline{\semantics{d_1}}$};
		\node [style=none] (40) at (-6.5, 0.25) {};
		\node [style=none] (41) at (-5.25, 0.25) {};
		\node [style=none] (42) at (-5.25, -1.75) {};
		\node [style=none] (43) at (-4, 0.25) {};
		\node [style=none] (44) at (-4, -1.75) {};
		\node [style=none] (45) at (-6.5, -1.75) {};
		\node [style=none] (46) at (3, 0.25) {$\overline{\semantics{d_1}}$};
		\node [style=none] (47) at (1.75, 1.25) {};
		\node [style=none] (48) at (3, 1.25) {};
		\node [style=none] (49) at (3, -0.75) {};
		\node [style=none] (50) at (4.25, 1.25) {};
		\node [style=none] (51) at (4.25, -0.75) {};
		\node [style=none] (52) at (1.75, -0.75) {};
	\end{pgfonlayer}
	\begin{pgfonlayer}{edgelayer}
		\draw [style=thick] (5.center) to (3.center);
		\draw [style=thick] (2.center) to (5.center);
		\draw [style=thick] (2.center) to (3.center);
		\draw [style=thick] (1.center) to (0.center);
		\draw [style=thick] (13.center) to (14.center);
		\draw [style=thick] (15.center) to (13.center);
		\draw [style=thick] (15.center) to (14.center);
		\draw [style=thick] (11.center) to (10.center);
		\draw [style=thick] (21.center) to (22.center);
		\draw [style=thick] (23.center) to (21.center);
		\draw [style=thick] (23.center) to (22.center);
		\draw [style=thick] (19.center) to (18.center);
		\draw (16.center) to (26.center);
		\draw (8.center) to (28.center);
		\draw [bend right=90, looseness=1.50] (32.center) to (30.center);
		\draw [bend left=90, looseness=1.75] (34.center) to (37.center);
		\draw [bend right=90, looseness=1.50] (33.center) to (26.center);
		\draw (38.center) to (30.center);
		\draw [bend right=75, looseness=0.75] (28.center) to (25.center);
		\draw [style=thick] (40.center) to (43.center);
		\draw [style=thick] (43.center) to (44.center);
		\draw [style=thick] (44.center) to (45.center);
		\draw [style=thick] (45.center) to (40.center);
		\draw [style=thick] (47.center) to (50.center);
		\draw [style=thick] (50.center) to (51.center);
		\draw [style=thick] (51.center) to (52.center);
		\draw [style=thick] (52.center) to (47.center);
		\draw (34.center) to (41.center);
		\draw (42.center) to (33.center);
		\draw (31.center) to (48.center);
		\draw (49.center) to (25.center);
	\end{pgfonlayer}
\end{tikzpicture}}
\endpgfgraphicnamed}
\end{center}

Here, since we are now doing two operations on the $\semantics{n_1}$ in reading (1)  and on the $\semantics{n_2}$ in reading (2),   we  need to apply  a $\mu$ to $\semantics{n_1}$    in reading (1)  and to $\semantics{n_2}$ in reading (2), right at the beginning. 

\section{Branching}

Branching of quantifiers happens  when there is a partial ordering on them. Henkin's prefix quantifiers are a form of branching.  In the sentence $d_1 n_1 v d_2 n_2$,  the partial order between two  quantifiers  is depicted as follows:
\begin{center}
 {%
\beginpgfgraphicnamed{Logic-Gen-Branching}
\begin{tikzpicture}
	\begin{pgfonlayer}{nodelayer}
		\node [style=none] (0) at (-2.5, 2.5) {$d_1 x \, n_1(x)$};
		\node [style=none] (1) at (-2.5, 0) {$d_2 y \, n_2(y)$};
		\node [style=none] (2) at (0.5, 1.25) {};
		\node [style=none] (3) at (-1, 2.5) {};
		\node [style=none] (4) at (-1, 0) {};
		\node [style=none] (5) at (1.75, 1.25) {$v(x,y)$};
	\end{pgfonlayer}
	\begin{pgfonlayer}{edgelayer}
		\draw (3.center) to (2.center);
		\draw (4.center) to (2.center);
	\end{pgfonlayer}
\end{tikzpicture}}
\endpgfgraphicnamed}
\end{center}

\noindent
When $d_1$ and $d_2$ have a linear ordering between them,  the above unfolds  to  the two readings discussed in the previous section.  The scope options seem to have resulted from having different  linear orderings  on quantifiers. This was recognised by  Hintikka \cite{Hintikka73} and Barwise \cite{Barwise79} who  showed  that branching  indeed happens  in natural language  and how quantifier cope ambiguities are manifestations of it.

The set-theoretic  semantics of   branching quantifiers is due to Barwise \cite{Barwise79} and it is as follows.  Suppose $\semantics{v} \subseteq \semantics{n_1} \times \semantics{n_2}$, then we have the following two cases:
\[
\begin{cases}
d_1, d_2 \  \mbox{upward monotone} & \exists X \subseteq \semantics{n_1}, \exists Y \subseteq \semantics{n_2}, d_1 \semantics{n_1}X \ \& \ d_2 \semantics{n_2} Y\ \& \ X \times Y \subseteq \semantics{v}\\
d_1, d_2 \  \mbox{downward monotone} & \exists X \subseteq \semantics{n_1}, \exists Y \subseteq \semantics{n_2}, d_1 \semantics{n_1}X \ \& \ d_2 \semantics{n_2}Y \ \& \ X \times Y \supseteq \semantics{v}
\end{cases}
\]

The general form of branching in our setting is the following left hand side diagram:

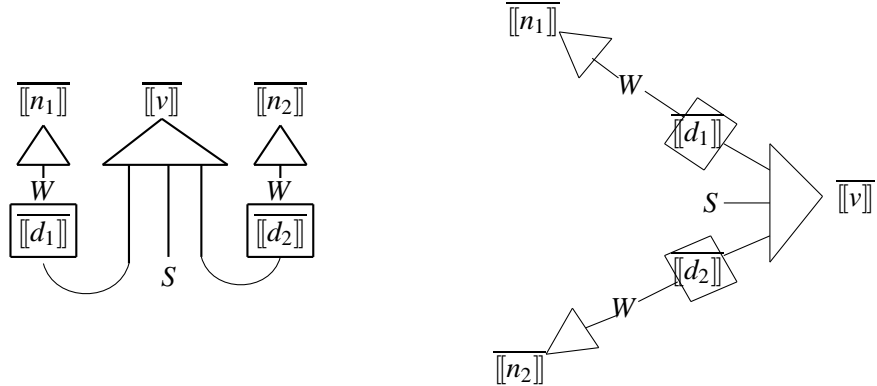
\begin{center}
  {%
\beginpgfgraphicnamed{Diag-Branch}
\begin{tikzpicture}[scale=0.7]
	\begin{pgfonlayer}{nodelayer}
		\node [style=none] (0) at (-7, 3.25) {};
		\node [style=none] (1) at (-7, 3.75) {};
		\node [style=none] (2) at (-7, 5.25) {};
		\node [style=none] (3) at (-6, 3.75) {};
		\node [style=none] (4) at (-7, 6.25) {$\ovl{\semantics{n_1}}$};
		\node [style=none] (5) at (-8, 3.75) {};
		\node [style=none] (6) at (-7, 2.75) {$W$};
		\node [style=none] (7) at (-7, 2.25) {};
		\node [style=none] (8) at (-7, 0) {};
		\node [style=none] (9) at (2, 6.25) {$\ovl{\semantics{n_2}}$};
		\node [style=none] (10) at (2, 2.25) {};
		\node [style=none] (11) at (2, 2.75) {$W$};
		\node [style=none] (12) at (2, 3.75) {};
		\node [style=none] (13) at (1, 3.75) {};
		\node [style=none] (14) at (2, 5.25) {};
		\node [style=none] (15) at (3, 3.75) {};
		\node [style=none] (16) at (2, 0.25) {};
		\node [style=none] (17) at (2, 3.25) {};
		\node [style=none] (18) at (0, 3.75) {};
		\node [style=none] (19) at (-3.75, 3.75) {};
		\node [style=none] (20) at (-3.75, 0) {};
		\node [style=none] (21) at (-4.75, 3.75) {};
		\node [style=none] (22) at (-2.5, 5.5) {};
		\node [style=none] (23) at (-2.5, 6.25) {$\ovl{\semantics{v}}$};
		\node [style=none] (24) at (-1, 3.75) {};
		\node [style=none] (25) at (-1, 0.25) {};
		\node [style=none] (26) at (-2.25, 3.75) {};
		\node [style=none] (27) at (-2.25, 0.25) {};
		\node [style=none] (28) at (-2.25, -0.5) {$S$};
		\node [style=none] (29) at (2, 0.25) {};
		\node [style=none] (30) at (2, 1.25) {$\overline{\semantics{d_2}}$};
		\node [style=none] (31) at (3.25, 2.25) {};
		\node [style=none] (32) at (3.25, 0.25) {};
		\node [style=none] (33) at (0.75, 0.25) {};
		\node [style=none] (34) at (0.75, 2.25) {};
		\node [style=none] (35) at (2, 2.25) {};
		\node [style=none] (36) at (-7, 0.25) {};
		\node [style=none] (37) at (-7, 1.25) {$\overline{\semantics{d_1}}$};
		\node [style=none] (38) at (-5.75, 2.25) {};
		\node [style=none] (39) at (-5.75, 0.25) {};
		\node [style=none] (40) at (-8.25, 0.25) {};
		\node [style=none] (41) at (-8.25, 2.25) {};
		\node [style=none] (42) at (-7, 2.25) {};
	\end{pgfonlayer}
	\begin{pgfonlayer}{edgelayer}
		\draw [style=thick] (5.center) to (3.center);
		\draw [style=thick] (2.center) to (5.center);
		\draw [style=thick] (2.center) to (3.center);
		\draw [style=thick] (1.center) to (0.center);
		\draw [style=thick] (13.center) to (15.center);
		\draw [style=thick] (14.center) to (13.center);
		\draw [style=thick] (14.center) to (15.center);
		\draw [style=thick] (12.center) to (17.center);
		\draw [style=thick] (21.center) to (18.center);
		\draw [style=thick] (22.center) to (21.center);
		\draw [style=thick] (22.center) to (18.center);
		\draw [style=thick] (19.center) to (20.center);
		\draw [style=thick] (24.center) to (25.center);
		\draw [style=thick] (26.center) to (27.center);
		\draw [bend right=75, looseness=1.25] (8.center) to (20.center);
		\draw [bend right=75, looseness=1.25] (25.center) to (16.center);
		\draw [style=thick] (34.center) to (31.center);
		\draw [style=thick] (31.center) to (32.center);
		\draw [style=thick] (32.center) to (33.center);
		\draw [style=thick] (33.center) to (34.center);
		\draw [style=thick] (41.center) to (38.center);
		\draw [style=thick] (38.center) to (39.center);
		\draw [style=thick] (39.center) to (40.center);
		\draw [style=thick] (40.center) to (41.center);
	\end{pgfonlayer}
\end{tikzpicture}}
\endpgfgraphicnamed}
  \hspace{2cm}
  {%
\beginpgfgraphicnamed{Diag-Branch-rotate}
\begin{tikzpicture}[scale=0.7]
	\begin{pgfonlayer}{nodelayer}
		\node [style=none] (0) at (-3.5, 6) {$\ovl{\semantics{n_1}}$};
		\node [style=none] (1) at (0.25, 3.5) {$W$};
		\node [style=none] (2) at (8.75, -0.75) {$\ovl{\semantics{v}}$};
		\node [style=none] (3) at (3.25, -1) {$S$};
		\node [style=none] (4) at (2.75, 1.75) {$\overline{\semantics{d_1}}$};
		\node [style=none] (5) at (5.5, 1.25) {};
		\node [style=none] (6) at (7.5, -0.75) {};
		\node [style=none] (7) at (5.5, -3.25) {};
		\node [style=none] (8) at (5.5, 0.25) {};
		\node [style=none] (9) at (5.5, -2.25) {};
		\node [style=none] (10) at (5.5, -1) {};
		\node [style=none] (11) at (3.75, 1.25) {};
		\node [style=none] (12) at (3.75, -3) {};
		\node [style=none] (13) at (3.75, -1) {};
		\node [style=none] (14) at (4.25, 2) {};
		\node [style=none] (15) at (3, 0.25) {};
		\node [style=none] (16) at (1.5, 1.25) {};
		\node [style=none] (17) at (2.75, 3) {};
		\node [style=none] (18) at (2.25, 2.25) {};
		\node [style=none] (19) at (0.75, 3.25) {};
		\node [style=none] (20) at (-1.75, 3.75) {};
		\node [style=none] (21) at (-0.5, 5.25) {};
		\node [style=none] (22) at (-2.5, 5.5) {};
		\node [style=none] (23) at (-1.25, 4.5) {};
		\node [style=none] (24) at (-0.25, 3.75) {};
		\node [style=none] (25) at (3.25, -2.25) {};
		\node [style=none] (26) at (4.25, -4) {};
		\node [style=none] (27) at (2.5, -4.75) {};
		\node [style=none] (28) at (1.5, -3) {};
		\node [style=none] (29) at (2, -4) {};
		\node [style=none] (30) at (0.5, -4.75) {};
		\node [style=none] (31) at (2.75, -3.5) {$\overline{\semantics{d_2}}$};
		\node [style=none] (32) at (0, -5) {$W$};
		\node [style=none] (33) at (-2, -5) {};
		\node [style=none] (34) at (-1, -6.5) {};
		\node [style=none] (35) at (-3, -6.75) {};
		\node [style=none] (36) at (-0.25, -5.25) {};
		\node [style=none] (37) at (-1.5, -5.75) {};
		\node [style=none] (38) at (-4, -7.25) {$\ovl{\semantics{n_2}}$};
	\end{pgfonlayer}
	\begin{pgfonlayer}{edgelayer}
		\draw (5.center) to (7.center);
		\draw (7.center) to (6.center);
		\draw (5.center) to (6.center);
		\draw (11.center) to (8.center);
		\draw (9.center) to (12.center);
		\draw (13.center) to (10.center);
		\draw (14.center) to (15.center);
		\draw (16.center) to (17.center);
		\draw (17.center) to (14.center);
		\draw (16.center) to (15.center);
		\draw (19.center) to (18.center);
		\draw (22.center) to (20.center);
		\draw (20.center) to (21.center);
		\draw (22.center) to (21.center);
		\draw (23.center) to (24.center);
		\draw (25.center) to (28.center);
		\draw (28.center) to (27.center);
		\draw (27.center) to (26.center);
		\draw (26.center) to (25.center);
		\draw (30.center) to (29.center);
		\draw (33.center) to (34.center);
		\draw (34.center) to (35.center);
		\draw (35.center) to (33.center);
		\draw (37.center) to (36.center);
	\end{pgfonlayer}
\end{tikzpicture}}
\endpgfgraphicnamed}
\end{center}

This diagram is very similar to the branching diagram. We have tried to make this apparent by rotating it at the right hand side  above.  This diagram on its own does not say much until we specify what is inside the $\overline{\semantics{v}}$ triangle. It is according to the content of $\overline{\semantics{v}}$ that the quantifiers interact with each other. We can encode Barwise's definition in the triangle when  defining concrete interpretations for $\overline{\semantics{v}}$ in any of the instantiations.  In the relational instantiation,  $\ovl{\semantics{v}}$ is an element of a  rank 3 tensors of  type ${\cal P}(U)  \otimes \{\star\} \otimes {\cal P}(U)$, and  $\ovl{\semantics{n_1}}$ and $\ovl{\semantics{n_2}}$ are elements of objects of type ${\cal P}(U)$.  Thus in the above diagram, $W$ will be instantiated as ${\cal P}(U)$ and $S$ as $\{\star\}$.   Given these,  we define:\\
\[
\begin{cases}
d_1, d_2 \  \mbox{upward monotone} &  \star \overline{\semantics{v}} (X, \star, Y) \iff \semantics{v}(X) \subseteq Y\\
d_1, d_2 \  \mbox{downward monotone} & \star \overline{\semantics{v}} (X, \star, Y) \iff \semantics{v} (X) \supseteq Y
\end{cases}
\]
where $ \semantics{v} ({X})$ is the forward image of $\semantics{X}$ in the binary relation $\semantics{v}$. We do not need to explicitly talk about $X \subseteq \semantics{n_1}$ and $Y \subseteq \semantics{n_2}$, since the lines of the diagram are identities on the object ${\cal P}(U)$, hence they carry subsets. When a line emanates from the triangle  that interprets $\ovl{\semantics{n_1}}$, it represents a subset of $\semantics{n_1}$, and when it emanates from the triangle that interprets $\ovl{\semantics{n_2}}$, it represents a subset of $\semantics{n_2}$.  Generalizing this definition to the abstract categorical framework is work in progress. 

\section{Conclusion}
In this paper, we reviewed our previous treatment of generalised quantifiers in the categorical compositional distributional semantics, where some objects have a bialgebra over them.  In this paper, we showed how one can deal with quantifier scope ambiguity in that setting. Scope ambiguity  gives rise to branching quantifiers.  We  showed how one may deal with branching in that setting as well.  A compositional  passage from  syntax to scope ambiguity and to branching  is  however missing from the current treatment; it constitutes future work.

\bibliographystyle{eptcs}
\bibliography{quant}

\begin{thebibliography}{10}
\providecommand{\bibitemdeclare}[2]{}
\providecommand{\surnamestart}{}
\providecommand{\surnameend}{}
\providecommand{\urlprefix}{Available at }
\providecommand{\url}[1]{\texttt{#1}}
\providecommand{\href}[2]{\texttt{#2}}
\providecommand{\urlalt}[2]{\href{#1}{#2}}
\providecommand{\doi}[1]{doi:\urlalt{http://dx.doi.org/#1}{#1}}
\providecommand{\bibinfo}[2]{#2}

\bibitemdeclare{article}{Barwise79}
\bibitem{Barwise79}
\bibinfo{author}{J.~\surnamestart Barwise\surnameend} (\bibinfo{year}{1979}):
  \emph{\bibinfo{title}{On Branching Quantifiers in English}}.
\newblock {\sl \bibinfo{journal}{Philosophical Logic}} \bibinfo{volume}{8}, pp.
  \bibinfo{pages}{47--80}, \doi{10.1007/BF00258419}.

\bibitemdeclare{article}{BarwiseCooper81}
\bibitem{BarwiseCooper81}
\bibinfo{author}{J.~\surnamestart Barwise\surnameend} \&
  \bibinfo{author}{R.~\surnamestart Cooper\surnameend} (\bibinfo{year}{1981}):
  \emph{\bibinfo{title}{Generalized quantifiers and natural language}}.
\newblock {\sl \bibinfo{journal}{Linguistics and Philosophy}}
  \bibinfo{volume}{4}, pp. \bibinfo{pages}{159--219}, \doi{10.1007/BF00350139}.

\bibitemdeclare{incollection}{Buszkowski-prg}
\bibitem{Buszkowski-prg}
\bibinfo{author}{Wojciech \surnamestart Buszkowski\surnameend}
  (\bibinfo{year}{2001}): \emph{\bibinfo{title}{Lambek Grammars Based on
  Pregroups}}.
\newblock In: {\sl \bibinfo{booktitle}{Logical Aspects of Computational
  Linguistics}}, {\sl \bibinfo{series}{Lecture Notes in Computer Science}}
  \bibinfo{volume}{2099}, \bibinfo{publisher}{Springer Berlin Heidelberg}, pp.
  \bibinfo{pages}{95--109}, \doi{10.1007/3-540-48199-0/6}.

\bibitemdeclare{article}{CoeckeSadrClark}
\bibitem{CoeckeSadrClark}
\bibinfo{author}{B.~\surnamestart Coecke\surnameend} \&
  \bibinfo{author}{M.~Sadrzadeh \surnamestart S.~Clark\surnameend}
  (\bibinfo{year}{2010}): \emph{\bibinfo{title}{Mathematical Foundations for a
  Compositional Distributional Model of Meaning}}.
\newblock {\sl \bibinfo{journal}{Lambek Festschirft, Linguistic Analysis, vol.
  36}} \bibinfo{volume}{36}, pp. \bibinfo{pages}{345--384}.

\bibitemdeclare{book}{Gardenfors87}
\bibitem{Gardenfors87}
\bibinfo{author}{P.~\surnamestart G\"{a}rdenfors\surnameend}
  (\bibinfo{year}{1987}): \emph{\bibinfo{title}{Generalized quantifiers
  linguistic and logical approaches}}.
\newblock \bibinfo{series}{Language Arts \& Disciplines},
  \bibinfo{publisher}{Springer Netherlands}, \doi{10.1007/978-94-009-3381-1}.

\bibitemdeclare{article}{HedgesSadr}
\bibitem{HedgesSadr}
\bibinfo{author}{J.~\surnamestart Hedges\surnameend} \&
  \bibinfo{author}{M.~\surnamestart Sadrzadeh\surnameend}
  (\bibinfo{year}{2016}): \emph{\bibinfo{title}{A Generalised Quantifier Theory
  of Natural Language in Categorical Compositional Distributional Semantics
  with Bialgebras}}.
\newblock {\sl \bibinfo{journal}{CoRR}} \bibinfo{volume}{abs/1602.01635}.
\newblock \urlprefix\url{http://arxiv.org/abs/1602.01635}.

\bibitemdeclare{article}{Hintikka73}
\bibitem{Hintikka73}
\bibinfo{author}{J.~\surnamestart Hintikka\surnameend} (\bibinfo{year}{1973}):
  \emph{\bibinfo{title}{Quantifiers vs. Quantification Theory}}.
\newblock {\sl \bibinfo{journal}{Dialectica}} \bibinfo{volume}{27}, pp.
  \bibinfo{pages}{329--358}, \doi{10.1111/j.1746-8361.1973.tb00624.x}.

\bibitemdeclare{inproceedings}{kartsaklis2012}
\bibitem{kartsaklis2012}
\bibinfo{author}{D.~\surnamestart Kartsaklis\surnameend},
  \bibinfo{author}{M.~\surnamestart Sadrzadeh\surnameend} \&
  \bibinfo{author}{S.~\surnamestart Pulman\surnameend} (\bibinfo{year}{2012}):
  \emph{\bibinfo{title}{A Unified Sentence Space for Categorical
  Distributional-Compositional Semantics: Theory and Experiments}}.
\newblock In: {\sl \bibinfo{booktitle}{Proceedings of 24th International
  Conference on Computational Linguistics (COLING 2012): Posters}},
  \bibinfo{address}{Mumbai, India}, pp. \bibinfo{pages}{549--558},
  \doi{10.1.1.360.2480}.

\bibitemdeclare{article}{McCurdy}
\bibitem{McCurdy}
\bibinfo{author}{M.~\surnamestart McCurdy\surnameend} (\bibinfo{year}{2012}):
  \emph{\bibinfo{title}{Graphical methods for {T}annaka duality of weak
  bialgebras and weak {H}opf algebras}}.
\newblock {\sl \bibinfo{journal}{Theory and applications of categories}}
  \bibinfo{volume}{26}(\bibinfo{number}{9}), pp. \bibinfo{pages}{233--280},
  \doi{10.1.1.300.4545}.

\bibitemdeclare{inproceedings}{PrellerSadr}
\bibitem{PrellerSadr}
\bibinfo{author}{A.~\surnamestart Preller\surnameend} \&
  \bibinfo{author}{M.~\surnamestart Sadrzadeh\surnameend}
  (\bibinfo{year}{2010}): \emph{\bibinfo{title}{Bell States and Negative
  Sentences in the Distributed Model of Meaning}}.
\newblock In \bibinfo{editor}{P.~Selinger \surnamestart B.~Coecke\surnameend,
  P.~Panangaden}, editor: {\sl \bibinfo{booktitle}{Electronic Notes in
  Theoretical Computer Science, Proceedings of the 6th QPL Workshop on Quantum
  Physics and Logic}}, \bibinfo{publisher}{University of Oxford}, pp.
  \bibinfo{pages}{141--153}, \doi{10.1016/j.entcs.2011.01.028}.

\bibitemdeclare{inproceedings}{RelPronMoL}
\bibitem{RelPronMoL}
\bibinfo{author}{M.~Sadrzadeh \surnamestart S.~Clark\surnameend, B.~Coecke}
  (\bibinfo{year}{2013}): \emph{\bibinfo{title}{The Frobenius Anatomy of
  Relative Pronouns}}.
\newblock In: {\sl \bibinfo{booktitle}{13th Meeting on Mathematics of Language
  (MoL)}}, pp. \bibinfo{pages}{41--51}.

\bibitemdeclare{incollection}{Westerstahl87}
\bibitem{Westerstahl87}
\bibinfo{author}{D.~\surnamestart Westerstahl\surnameend}
  (\bibinfo{year}{1987}): \emph{\bibinfo{title}{Branching Generalized
  Quantifiers and Natural Language}}.
\newblock In \bibinfo{editor}{P.~\surnamestart G\"ardenfors\surnameend},
  editor: {\sl \bibinfo{booktitle}{Generalized Quantifiers: Linguistic and
  Logical Approaches}}, \bibinfo{publisher}{Reidel},
  \bibinfo{address}{Dordrecht}, pp. \bibinfo{pages}{269--298},
  \doi{10.1007/978-94-009-3381-1/10}.

\end{thebibliography}
\end{document}